\documentclass[iicol]{sn-jnl}

\usepackage{graphicx}%
\usepackage{multirow}%
\usepackage{amsmath,amssymb,amsfonts}%
\usepackage{amsthm}%
\usepackage{mathrsfs}%
\usepackage[title]{appendix}%
\usepackage{xcolor}%
\usepackage{textcomp}%
\usepackage{manyfoot}%
\usepackage{booktabs}%
\usepackage{algorithm}%
\usepackage{algorithmicx}%
\usepackage{algpseudocode}%
\usepackage{listings}%

\usepackage{lipsum} 
\usepackage{booktabs} 
\usepackage{tabularx} 
\usepackage{soul}

\begin{document}

\title[Article Title]{ENet-21: An Optimized light CNN Structure for Lane Detection}
\author*[1]{\fnm{Seyed Rasoul} \sur{Hosseini}}\email{rasoul.hosseini7@gmail.com}

\author[1]{\fnm{Hamid} \sur{Taheri}}\email{taheri.hamiid@gmail.com}

\author[1]{\fnm{Mohammad} \sur{Teshnehlab}}\email{teshnehlab@eetd.kntu.ac.ir}

\affil[1]{\orgdiv{Intelligent Systems Laboratory (ISLAB), Faculty of Electrical \& Computer Engineering}, \orgname{K.N. Toosi University of Technology}, \city{ Tehran}, \country{Iran}}

\abstract{Lane detection for autonomous vehicles is an important concept, yet it is a challenging issue of driver assistance systems in modern vehicles. The emergence of deep learning leads to significant progress in self-driving cars. Conventional deep learning-based methods handle lane detection problems as a binary segmentation task and determine whether a pixel belongs to a line. These methods rely on the assumption of a fixed number of lanes, which does not always work. This study aims to develop an optimal structure for the lane detection problem, offering a promising solution for driver assistance features in modern vehicles by utilizing a machine learning method consisting of binary segmentation and Affinity Fields that can manage varying numbers of lanes and lane change scenarios. In this approach, the Convolutional Neural Network (CNN), is selected as a feature extractor, and the final output is obtained through clustering of the semantic segmentation and Affinity Field outputs. Our method uses less complex CNN architecture than existing ones. Experiments on the TuSimple dataset support the effectiveness of the proposed method.
}



\maketitle
\section{Introduction} \label{sec1}

Researchers' main interest in computer vision for academia and industry is autonomous vehicles. Keeping the car between lane lines, referred to as Lane Detection, is critical for car safety. Lane detection has considerably improved through the adoption of deep learning \cite{b1}.

According to \cite{b1}, there are two primary methods for lane detection: classical algorithms and machine learning-based algorithms. Considering \cite{b1}, the former was derived from handmade feature selection, and the major issue was the reliance on the prior knowledge and perceived features of the problem. Therefore, final results are achieved deterministically following pixel detailing on the extracted features. These algorithms depend on the input images and their quality rather than machine learning-based methods, which should contain the desired features with high visibility. Due to prior feature selection, the input image can be divided into different regions of interest, resulting in lighter and faster detailing.

The advancement of traditional methods is related to two major areas: system reliability and improved understanding of the environment.  Since the proposed system is mainly based on computer vision approaches, reliability is vital but isn't easy to maintain. All feasible noises and environmental conditions may influence the system’s performance. To resolve this problem, methods should be general, without any bold presumption about what scheme the system has to perform. In addition, numerous methods need to be weighted in parallel or applied in series and dynamically change to the best approach for each circumstance when the presumptions of the prior algorithm are not met \cite{b1}.

Over the last few years, a new approach to the problem of lane detection has been presented concerning classical algorithms. Rather than operating at a low level with pixels and depending on prior knowledge of noticeable features of lane lines and their shape, a machine learning-based algorithm has been suggested \cite{b2}.
Machine learning-based algorithms can perform automatic feature extraction, simplify the process, and provide a more robust solution, so they are more efficient against fluctuating elements and variable environments. The most common structures for the lane detection problem are Convolutional Neural Networks (CNNs).
CNN is especially suited to this problem because the architecture of the convolutional layer allows the simultaneous examination of several different image features for lane structure detection. Furthermore, the training enables us to automatically select and tune characteristics by assigning weights and optimizing the selection in complicated scenarios. Therefore, simplifying the task into an image binarization and giving a label to each image pixel \cite{b1}.

In recent studies, end-to-end approaches have been implemented, which means the whole procedure is performed by CNN, using the input images, without any initial preprocessing. For instance, in \cite{b2}, CNNs are applied to highway images to detect lanes and obstacles, and in \cite{b3}, an end-to-end algorithm called Dual-View CNN was implemented on front and top views for detection.
Another study \cite{b4} suggested a Spatial Convolutional Neural Network (SCNN) to evade performance degradation caused by obstacles blocking lane markings. In this structure, in addition to the communication between the neurons of different layers, communication is also established between the neurons of the same layer. This additional level of communication allows us to reach smoothness in the predicted lanes without interruption because of obstacles.

Another difficulty is the need to detect multiple lanes in one image. The first practical solution to this problem is to model the structure to behave as a classifier with multi-class where every lane is attributed to a class. In \cite{b5}, the proposed system is a multi-task CNN, which performs both the segmentation and the embedding branches. The segmentation branch performs as a binary segmentation on the input image. The embedding branch continues the previous stage, assigns a lane identification number to each lane pixel, and specifies the different lanes.

However, although a deep learning-based approach has advantages, some difficulties must be handled. The main issue is caused by the complication of the CNN structure, which may conflict with the real-time expansion. Here, the computational cost is considered, which can be handled by running the algorithm on suitable hardware or reduced by lighting the network.

The contributions of this study are as follows: 1) We present a light convolutional neural network (CNN) backbone with fewer FLOPs and parameters. We can achieve almost the same performance as existing architectures proposed for lane detection. 2) We use affinity fields for segmentation that can effectively group and link pixels that belong to shapeless entities, such as lane markings. 3) Our proposed method is comparable to or surpasses the leading techniques in the TuSimple benchmark dataset without complicated post-processing.

In this study, a machine learning-based method concerning CNN was employed for feature extraction on the TuSimple dataset to tackle the mentioned issue. Also, vector fields called affinity fields are used for clustering and assigning pixels to each pixel. Subsequently, we propose a light CNN structure for lane detection problems.

\section{Related Works}\label{sec2}

In recent years, significant improvements have been made regarding lane detection problems. In the previous part, we introduced different approaches to lane detection, including traditional and deep learning-based methods. In this section, we aim to review the latest research on the problem.

Lane detection aims to obtain the accurate shape of lanes and differentiate them from each other. According to lane modeling, we can divide current deep learning-based methods into several categories \cite{b6}. Semantic and instance segmentation-based \cite{b4}, \cite{b7}, \cite{b8}, \cite{b9}, key point detection \cite{b10}, \cite{b11}, gridding \cite{b12}, \cite{b13}, \cite{b14}, Ploynominal \cite{b15}, \cite{b16}, and Anchor-based \cite{b17}.

Focus on Local in \cite{b11}, inspired by the algorithm designed for human body key point detection, to recognize the critical points of the road lanes, and uses local geometry construction to rectify the location of the key point pixels.

PINet \cite{b10} benefits from Stacked Hourglass Network as a key point detection method. In this method, predicting the offset in the x and y directions considering an L2 regularization loss for clustering lanes helps to achieve better accuracy for indicated points.

CondLaneNet’s CNN \cite{b12} is designed based on the original conditional instance segmentation technique and row anchor formulation. It includes proposal and conditional shape heads. The Proposal head generates instance segmentation for lanes, and the conditional shape head produces the point position in the lane. Then, the row anchor-based lane detection was performed. However, recognizing start points is problematic in some complicated scenarios, which leads to almost substandard performance. Comparing CondLaneNet to other methods, such as FOLOLane and RESA, it does not have the best performance, and there is a lack of global spatial information on the TuSimple dataset. The reason is that images of the TuSimple dataset have more structured information than other datasets, such as CULane.

SCNN \cite{b4} considers lane detection problems a multiclass semantic segmentation. It is a spatial CNN module that recurrently aggregates spatial information to complete the discontinuous segmentation predictions and decodes the output segmentation by performing heuristic post-processing. Hence, it is not real-time, has low inference speed, and only struggles to solve this issue after optimizing the Zheng et al. method \cite{b10}. Other approaches explore knowledge distillation \cite{b19} or generative modeling \cite{b20} but result in better performance of the seminal SCNN. Moreover, these methods mainly consider a defined number (e.g., 4) of lanes. LaneNet \cite{b5} uses an instance segmentation pipeline to handle a variable number of lanes. However, generating lane instances requires post-inference clustering.

Line-CNN \cite{b21} benefits from classic Faster R-CNN \cite{b22} as an end-to-end lane detector, but it struggles to be real-time with high latency ($<$30 FPS). LaneATT \cite{b17} presents a more common end-to-end detection technique that results in better performance. It introduces a novel anchor-based attention technique that aggregates global information. Also, it manages to achieve good performance and gain both high adequacy and efficiency.

\section{Methodology}\label{sec3}

As reviewed in the previous section, various approaches were represented to solve the lane marker detection problem, each with pros and cons. However, our methodology trains an end-to-end convolutional neural network for lane prediction in such a way that it deals with the restriction on the numerous lanes mentioned. We can solve this problem by considering lane prediction as an example of a segmentation problem. With inspiration from \cite{b7}, we train the model to predict semantic segmentation labels and pixel-wise affinity fields. These affinity fields are divided horizontally and vertically (HAF and VAF, respectively). They map the 2D locations of the input image into a 2D unit vector, hence can be assumed as vector fields. In the vertical affinity field, this unit vector encodes a lane pixel vector direction toward the next lane pixels of the above row.

\begin{figure}
    \centering
    \includegraphics[width=0.49\textwidth]{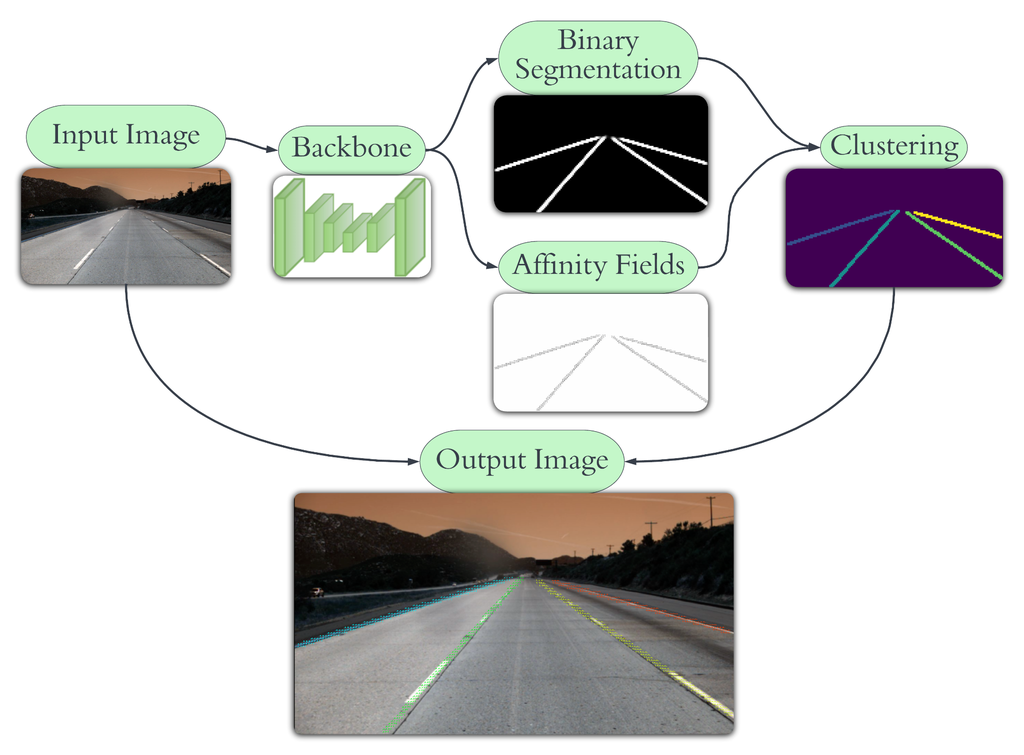}
    \caption{The methodology for lane marker detection.}\label{f1}
\end{figure}

On the contrary, the process is a bit different in the horizontal affinity field. The unit vector points toward the center pixel of the lane in the current row, which enables us to cluster lanes with different widths. In a postprocessing step, the VAF and HAF, together with the semantic lane segmentation, make it possible to generate lanes through foreground pixel clustering. We illustrate this method in Fig. \ref{f1}. Alongside this approach, the main focus is on CNN architecture, called backbone. The effort is to obtain a lighter structure than the existing ones with proper performance. In the following subsections, we will demonstrate the proposed architecture for binary segmentation and affinity fields and its training and inference procedure.

\subsection{Architecture}

The latest lane detection methods make use of different kinds of backbone structures, but their most well-known are the ResNet family \cite{b23}, ENet \cite{b24}, and ERFNet \cite{b25}. Even though the mentioned architectures have shown profit across various tasks since lane detection is a real-time problem, we believe that the number of parameters can be reduced for backbones. Table \ref{t1} illustrates the trainable parameters for mentioned architectures. For this purpose, we use the ENet backbone presented in \cite{b24} to obtain a lighter structure.

\begin{table*}[ht!]
  \begin{center}
    \caption{Different CNN architecture specifications.}
    \label{t1}
    \begin{tabular}{ccc} 
      \textbf{Structure} & \textbf{Number of Layers} &              \textbf{Number of Parameters}\\
       \hline \hline
        \multicolumn{1}{l|}{      Resnet-18 \cite{b23}} & 18 & 11.17M\\
        \multicolumn{1}{l|}{      Resnet-34 \cite{b23}} & 34 & 21.28M\\
        \multicolumn{1}{l|}{      Resnet-101 \cite{b23}} & 101 & 42.51M\\
        \multicolumn{1}{l|}{      ERFNet \cite{b25}} & 23 & 2.1M\\
        \multicolumn{1}{l|}{      ENet \cite{b24}} & 29 & 0.37M\\
            
    \end{tabular}
  \end{center}
\end{table*}

ENet (Efficient Neural Network) can operate as per-pixel binary segmentation in real-time. It’s also faster, has fewer FLOPs with fewer parameters, and yields similar or, in some cases, better accuracy than current models.

Table \ref{t2} depicts the structure of our model. It’s divided into several stages, like reference structure, as horizontal lines in Table \ref{t2} separate them. We report output sizes based on resized input images of TuSimple with a resolution of $(640 \times 352)$ and three filters. Here, just like ResNets \cite{b23}, that considers having a single main branch alongside extensions of convolutional filters that separate from it and then sum up with an element-wise addition. Each block contains three convolutional layers: a first projection that reduces the input dimensions, a main convolutional layer, and an expansion layer. We place the Batch Normalization layer \cite{b26} and PReLU activation function \cite{b27} after each convolutional layer. As in \cite{b24}, we call them bottleneck modules. When we consider the bottleneck as a downsampler, in addition to the mentioned structure, a max pooling layer is considered in the main branch.

The first projection is also changed with a convolution with a kernel size of $(2 \times 2)$ and stride 2 in both dimensions. We consider padding with zero value to match feature map numbers. The convolution layer is either a regular, dilated, or full convolution (known as deconvolution) with $(3 \times 3)$ filters. To consider a regularization, we apply Spatial Dropout [28]. The initial stage consists of a single block. Stage 1 contains three bottleneck blocks, while stages $2$ and $3$ have the same architecture, except that stage 3 does not have the downsampling bottleneck at the beginning. We call the first three steps the encoder. Stages $4$ and $5$ are called the decoder.

We did not consider bias terms for any of the projections to decrease the number of parameters for kernel and overall memory processing, as cuDNN \cite{b29} uses separate kernels for convolution and bias. This didn’t change the accuracy. We use Batch Normalization \cite{b26} after each convolutional layer and following nonlinearity. We use max-unpooling instead of max-pooling for decoding structure, and padding is replaced with spatial convolution without bias. Since the initial block acted on the three channels of the input frame, while the final output has C feature maps, we did not use pooling indices in the last upsampling module. In addition, for implementation reasons, we considered only a simple full convolution as the last module of the network, which alone takes up a sizeable share of the decoder acting time.

Also, in stage $5$, we have three parallel heads to create binary segmentation and affinity field output. These $3$ contain two sampling blocks and a simple $(1 \times 1)$ convolution to adjust dimensions shown in Layers $19-21$. Variable C is different for binary segmentation and affinity field output.

\begin{table}[ht]
\caption{ENet architecture. Output sizes are given
for input of ($640\times352\times3$).}
\label{t2}
\begin{tabular}{ccccc}
                              &    & \textbf{CNN Layer}         & \textbf{Type}         & \textbf{Output size} \\ \hline \hline 
    \multirow{15}{*}{\rotatebox[origin=c]{90}{Encoder}} & 1  & initial       &              & $320\times176\times16$ \vspace{1mm}  \\ 
                              & 2  & bottleneck1.0 & downsampling & $160\times88\times64$   \\
                              & 3  & bottleneck1.1 & dilated 2    & $160\times88\times64$   \\
                              & 4  & bottleneck1.2 & dilated4     & $160\times88\times64$  \vspace{1mm} \\ 
                              & 5  & bottleneck2.0 & downsampling & $88\times44\times128$   \\
                              & 6  & bottleneck2.1 &              & $88\times44\times128$   \\
                              & 7  & bottleneck2.2 & dilated 2    & $88\times44\times128$   \\
                              & 8  & bottleneck2.3 & dilated 4    & $88\times44\times128$   \\
                              & 9  & bottleneck2.4 & dilated 8    & $88\times44\times128$   \\
                              & 10 & bottleneck2.5 & dilated 16   & $88\times44\times128$  \vspace{1mm} \\  
                              & 11 & bottleneck3.1 &              & $88\times44\times128$   \\
                              & 12 & bottleneck3.2 & dilated 2    & $88\times44\times128$   \\
                              & 13 & bottleneck3.3 & dilated 4    & $88\times44\times128$   \\
                              & 14 & bottleneck3.4 & dilated 8    & $88\times44\times128$   \\
                              & 15 & bottleneck3.5 & dilated 16   & $88\times44\times128$  \vspace{0.35mm} \\ \hline 
    \multirow{6}{*}{\rotatebox[origin=c]{90}{Decoder}}         & 16 & bottleneck4.0 & upsampling   & $160\times88\times64$   \\
                              & 17 & bottleneck4.1 &              & $160\times88\times64$   \\
                              & 18 & bottleneck4.2 &              & $160\times88\times64$  \vspace{1mm} \\  
                              & 19 & bottleneck5.0 &              & $160\times88\times64$   \\
                              & 20 & bottleneck5.1 &              & $160\times88\times64$   \\
                              & 21 & conv          &              & $160\times88\times$C
\end{tabular}
\end{table}

\subsection{Affinity Fields}
As well as the binary segmentation head, our model is designed to train and predict affinity fields given in \cite{b7}. We can consider the HAF and VAF as $\overrightarrow{H}\left( ., . \right)$ and $\overrightarrow{V}\left( ., . \right)$ for each input image that assigns a unit vector to coordinates of $\left( x, y \right)$ in a frame. As mentioned, the HAF and VAF enable the model to cluster lane pixels horizontally and vertically. After predicting the binary segmentation mask and affinity fields, a straightforward decoding process was performed row-by-row from bottom to top to cluster lane pixels. Following this subsection, the process of using the ground truth to create vector fields and applying the predicted vector fields to decode lanes instances has been described.

\textbf{Creating HAFs and VAFs:} Affinity fields are generated using ground truth masks obtained from the dataset annotations, as in \eqref{eq1} and \eqref{eq2}. This process is done row-by-row and executes from bottom to top. The rows are selected based on the y coordinates of ground truth. For each row y in a frame, the ground truth is used to compute the HAF vectors for each lane point  $\left( x_i^l, y \right)$ vector field mapping $\overrightarrow{H}_{gt}\left( ., . \right)$ as follows:

\begin{equation}
\begin{aligned}
    \overrightarrow{H}_{gt}\left( x_i^l, y \right) = \left(\frac{\overline{x}_y^l - {x}_i^l}{\left|\overline{x}_y^l - {x}_i^l \right|}, \frac{ y-y}{\left| y-y \right|} \right)^T \\
    = \left(\frac{\overline{x}_y^l - {x}_i^l}{\left|\overline{x}_y^l - {x}_i^l \right|},0 \right)^T,
\label{eq1}
\end{aligned}
\end{equation}

where $i$ in $x_i^l$ denotes index for different x-coordiantes of lane $l$ and $\overline{x}_y^l$ is the center of lane $l$ in row $y$. Similarly, using the ground truth to compute the VAF vectors for each lane point $\left( x_i^l, y \right)$ in row $y$ vector field mapping $\overrightarrow{V}_{gt}\left( ., . \right)$ as follows:

\begin{equation}
\begin{aligned}
    \overrightarrow{V}_{gt}\left( x_i^l, y \right) = \left(\frac{\overline{x}_{y-1}^l - {x}_i^l}{\left|\overline{x}_{y-1}^l - {x}_i^l \right|}, \frac{ y-1-y}{\left| y-1-y \right|} \right)^T \\
    = \left(\frac{\overline{x}_{y-1}^l - {x}_i^l}{\left|\overline{x}_{y-1}^l - {x}_i^l \right|},-1 \right)^T,
\label{eq2}
\end{aligned}
\end{equation}

where  $\overline{x}_{y-1}^l$ is the center of lane $l$ in row $y-1$. It should be mentioned that, different from the HAF, unit vectors in the VAF point to the center of the lane in the previous row.

\textbf{Decoding HAFs and VAFs:} After training the structure to create the HAFs and VAFs described earlier, a decoding process is performed during testing to cluster foreground pixels into lanes. This procedure similarly performs bottom-to-top row-by-row.

Let’s assume vector field $\overrightarrow{H}_{pred}\left( ., . \right)$ belongs to the predicted HAF, based on \eqref{eq3} first assign to cluster foreground pixels in a row $y-1$ as follows:

\begin{equation}
\begin{aligned}
    \footnotesize{{c}_{haf}^*\left( {x_i^{fg}}, {y-1} \right) =  \begin{cases}
    C^{k+1} & \text{if} \overrightarrow{H}_{pred}\left( {x_{i-1}^{fg}}, {y-1} \right)_0 \leq 0\\
    &\wedge  \overrightarrow{H}_{pred}\left( {x_{i}^{fg}}, {y-1} \right)_0 > 0,\\
    C^{k}              & \text{otherwise}.
\end{cases}
}
\label{eq3}
\end{aligned}
\end{equation}

Here ${c}_{haf}^*\left( {x_i^{fg}}, {y-1} \right)$ stands for the optimal cluster assignment for a foreground pixel $\left( {x_i^{fg}}, {y-1} \right)$; $C^k$ and $C^{k+1}$ represent divergent clusters indexed by $k$ and $k+1$, respectively.

Then, by using the vector field $\overrightarrow{V}_{pred}\left( ., . \right)$ belonging to the VAF, the predicted horizontal clusters from \eqref{eq3} are assigned to existing lanes indexed by $l$ as follows:

\begin{equation}
\begin{aligned}
    c_{vaf}^*(l)=\arg\min_{C^k} d^{C^k}(l).
\label{eq4}
\end{aligned}
\end{equation}

where

\begin{equation}
\begin{aligned}
    d^*(l)=\min_{C^k} d^{C^k}(l).
\label{eq5}
\end{aligned}
\end{equation}

Here, ${d}^{C^k}(l)$ in \eqref{eq5} denotes the error of associating cluster ${C^k}$ to an existing lane $l$:

\begin{equation}
\begin{aligned}
    d^{C^k}(l)=\frac{1}{N_y^l} \sum_{i=0}^{N_y^l-1} \Bigg\| \left( \overline{x}^{C^k}, y-1 \right)^T - \left( x_i^l, y \right)^T \\ - \overrightarrow{V}_{pred}\left( x_i^l, y \right)\cdot \big\| \left( \overline{x}^{C^k}, y-1 \right)^T - \left( x_i^l, y \right)^T \big\| \Bigg\|.
\label{eq6}
\end{aligned}
\end{equation}

where $N_y^l$ is the number of pixels corresponding to lane $l$ in row $y$. Repeating the mentioned procedures row-by-row, every foreground pixel can be assigned to its respective lanes from the bottom to the top.

\subsection{Loss Functions}
We train the proposed model with a separate loss function consisting of three parts at each prediction head. First, we applied weighted binary cross-entropy (WBCE) loss for the semantic segmentation head, which is an ideal loss for unbalanced semantic segmentation tasks. The raw logits produced by the model are passed through a sigmoid activation for normalization and then applied to WBCE loss. The loss is defined as follows: 

\begin{equation}
\begin{aligned}
    L_{WBCE} = -\frac{1}{N} \sum_i \biggr[ w\cdot t_i \cdot \log{(o_i)} + \left(1 - t_i \right) \\
    \cdot \log{(1 - o_i)} \biggr],
\label{eq7}
\end{aligned}
\end{equation}

where $t_i$ demonstrate target value belonging to pixel $i$ and $o_i$ is the sigmoid output of the pixel. Since the lane detection problem is an imbalanced segmentation task, to change penalization for foreground pixels, a weight $w$ was considered. Another loss was considered when computing the unbalanced dataset further. Hence, Intersection over Union loss was used for the segmentation head:

\begin{equation}
\begin{aligned}
    L_{IoU} = \frac{1}{N} \sum_i \biggr[ 1 - \frac{t_i \cap o_i}{t_i \cup o_i} \biggr].
\label{eq8}
\end{aligned}
\end{equation}

In addition, a simple L1 regression loss was applied only to the foreground locations of both the vertical and horizontal affinity fields heads of the model as follows:

\begin{equation}
\begin{aligned}
    L_{AF}  = \frac{1}{N_{fg}} \sum_i \biggr[ \left| t_i^{haf} - o_i^{haf} \right| + \left| t_i^{vaf} - o_i^{vaf} \right| \biggr].
\label{eq9}
\end{aligned}
\end{equation}

The general loss function applied to the model is the summation of the individual-mentioned losses:

\begin{equation}
\begin{aligned}
    L_{general} = L_{WBCE} + L_{IoU} + L_{AF}
\label{eq10}
\end{aligned}
\end{equation}

\section{Experimental Results}\label{sec4}

\subsection{Dataset}
Currently, the TuSimple dataset \cite{b18} is a substantial and widely used dataset for evaluating our approach to the lane detection problem. It includes $6408$ videos from US highways in good and moderate weather conditions, divided into three sets: training, validation, and testing. The resolution of the images is $1280 \times 720$. They were collected at different hours with variable numbers of lanes. For each video, $20$ frames have been provided, and there are $128160$ frames in total. However, only the last frame in each video is annotated so there are $6408$ images available. The annotations’ format is .json, containing the x-position of the lanes at several discretized y-positions. Also, in this dataset, the ego and left and right lanes are annotated for each image. In lane-changing scenarios, a 5th lane is considered in frames to avoid confusion.

\subsection{Metrics}
To compare our approach with other existing methods for lane detection problems, we used the general evaluation metrics for the TuSimple dataset. We consider the official metric presented in \cite{b18} as the evaluation benchmark. Based on this benchmark, accuracy is determined as the correct average number of vertices in each image:

\begin{equation}
\begin{aligned}
    Accuracy = \frac{N_{correct}}{N_{gt}},  
\label{eq11}
\end{aligned}
\end{equation}

$N_{correct}$ is the number of lane vertices predicted correctly, and $N_{gt}$ is the total number of ground truth lane vertices in the annotation. Also, according to the benchmark,  the false positive (FP) and false negative (FN) rates are reported as follows:

\begin{equation}
\begin{aligned}
    FP = \frac{N_{false}}{N_{pred}},
    \label{eq12}
    \end{aligned}
    \end{equation}
    \begin{equation}
    \begin{aligned}
    FN = \frac{N_{missed}}{N_{gt}},
\label{eq13}
\end{aligned}
\end{equation}

Where $N_{false}$ is the number of wrong predicted lanes, $N_{pred}$ is the number of all predicted lanes, and $N_{missed}$ is the number of missed ground-truth lanes in the predictions.

Furthermore, according to the obtained values of Accuracy, FP, and FN and considering $TP = accuracy$, we report the F1-score as follows:
\begin{equation}
\begin{aligned}
    F1-score = 2 \times \frac{Precision \times Recall}{Precision+Recall},
    \label{eq14}
\end{aligned}
\end{equation}
where $Precision = \frac{TP}{TP+FP}$ and $Recall = \frac{TP}{TP+FN}$.

\subsection{Implementation}
At first, we resized the input images of the TuSimple dataset to half of their original width and length (i.e., 640$\times$352) and, in addition, reshaped the segmentation masks and ground truth to one-eighth of the initial resolution based on the network’s downsizing. Since we process smaller images and pixels, this has the advantage of faster processing. The processing time relies on the lane number in each image, the model’s output quality, and the output size. To assign a lane marker mask for each frame, we generate them concerning .json labels, which are x and y coordinates. 

For optimization, we used the Adam optimizer with a scheduled learning rate and initial value of 0.0005, which changes every ten epochs, with a weight decay of 0.01. Mini-batch has applied with eight images per GPU and a maximum of 40 epochs for training on Tusimple. To reduce overfitting, we applied Dropout after every bottleneck with a probability of 0.2. In addition, to preserve the parameters of the best-performed model on the validation set, we considered early stopping. Furthermore, we also applied data augmentation during training, like crops, random rotations, horizontal flips, and scales.
\subsection{Results}
We have made extensive comparisons with other existing state-of-the-art architectures to examine the validity of our method and structure. We considered multiple backbones, i.e., the ResNet family \cite{b23} and ERFNet \cite{b25}, which have used different methods. As illustrated in Table \ref{t3}, the suggested architecture seized competitive results on the TuSimple dataset. Compared to current performances, a notable achievement is the f1-score and small False Positive ratio, achieved without complicated post-processing. Interestingly, a more complex network appears to have lower f1-score results, e.g., R-34-E2E \cite{b14}. The explanation is that the number of TuSimple training images is insufficient, and networks should deal with overfitting, so a much lighter structure is needed to solve this issue. In addition, the false positive rate in our approach is the second lowest (0.0242) among the current results.
Furthermore, as demonstrated in Table \ref{t4}, with only 0.25 million parameters and 3.14G FLOPs, our model is much lighter and faster than commonly used models. This means our technique rarely detects a pixel incorrectly as a lane compared to other methods and ensures confident lane pixel detection. While acquiring higher results of f1-score to other methods which use ENet as their backbone architectures, such as ENet-SAD \cite{b19} and ENet-Label \cite{b30}, our approach has slightly lower performance compared to current state-of-the-art methods such as CLRNet \cite{b31}, PINet \cite{b10}, and R34-ATT \cite{b17}. However, the reason for our high false negative rate compared to others is the incorrect detection of lane pixels at the ends of each lane.

\begin{table*}[ht]
\centering
\caption{Comparison of different methods and structures on the TuSimple test
set.}
\label{t3}
\begin{tabular}{c c c c c}
    \textbf{Method}                                   & \textbf{Accuracy} (\%) & \textbf{FP}     & \textbf{FN}     & \textbf{F1-score} (\%)  \\ \hline \hline
    \multicolumn{1}{l|}{ENet-SAD \cite{b19}}            & 96.64         & 0.0602 & 0.0205 & 95.92                        \\
    \multicolumn{1}{l|}{ENet-Label \cite{b30}}          & 96.29         & 0.0602 & 0.0205 & 95.23                        \\
    \multicolumn{1}{l|}{ERF-E2E \cite{b14}}             & 96.02         & 0.0722 & 0.0218 & 96.25                        \\
    \multicolumn{1}{l|}{DLA34-AF \cite{b7}}            & 95.61         & 0.0280 & 0.0418 & 96.48                        \\
    \multicolumn{1}{l|}{R34-ATT \cite{b17}}             & 95.63         & 0.0353 & 0.0292 & 96.77                        \\
    \multicolumn{1}{l|}{ERF-FOLO \cite{b11}}            & \textbf{96.92}         & 0.0447 & 0.0228 & 96.63                        \\
    \multicolumn{1}{l|}{R34-E2E \cite{b14}}             & 96.22         & 0.0308 & 0.0376 & 96.58                        \\
    \multicolumn{1}{l|}{CLRNet \cite{b31}}              & 96.84         & \textbf{0.0228} & \textbf{0.0192} & \textbf{97.89}                        \\
    \multicolumn{1}{l|}{PINet \cite{b10}}               & 96.75         & 0.0310 & 0.0250 & 97.20                        \\ \hline
    \multicolumn{1}{l|}{\textit{ENet(Ours)}}          & \textit{95.88}         & \textit{0.0268} & \textit{0.0389} & \textit{96.68}                   
\end{tabular}

\end{table*}

\begin{table*}[ht]
  \begin{center}
    \caption{Comparison of different CNN architecture parameters.}
    \label{t4}
    \begin{tabular}{ccc} 
      \textbf{Method} & \textbf{FLOPs} & \textbf{Number of Parameters}\\
      \hline \hline
    \multicolumn{1}{l|}{ENet-SAD \cite{b19}} & 4.2G & 0.37M\\
    \multicolumn{1}{l|}{ENet-Label \cite{b30}} & 4.2G & 0.37M\\
    \multicolumn{1}{l|}{ERF-E2E \cite{b14}} & 41.66G & 2.71M\\
    \multicolumn{1}{l|}{DLA34-AF \cite{b7}} & 44.58G & 20.17M\\
    \multicolumn{1}{l|}{R34-ATT \cite{b17}} & 33G & 21.8M\\
    \multicolumn{1}{l|}{ERF-FOLO \cite{b11}} & 41.66G & 2.1M\\
    \multicolumn{1}{l|}{R34-E2E \cite{b14}} & 33G & 21.8M\\
    \multicolumn{1}{l|}{CLRNet \cite{b31}} & 33G & 21.8M\\
    \multicolumn{1}{l|}{PINet \cite{b10}} & - & 4.06M\\ \hline
    \multicolumn{1}{l|}{\textit{ENet (ours)} }& \textit{\textbf{3.14G}} & \textit{\textbf{0.25}}M\\

    \end{tabular}
  \end{center}
\end{table*}

\begin{figure}[!ht]
    \centering
    \includegraphics[width=2.4cm]{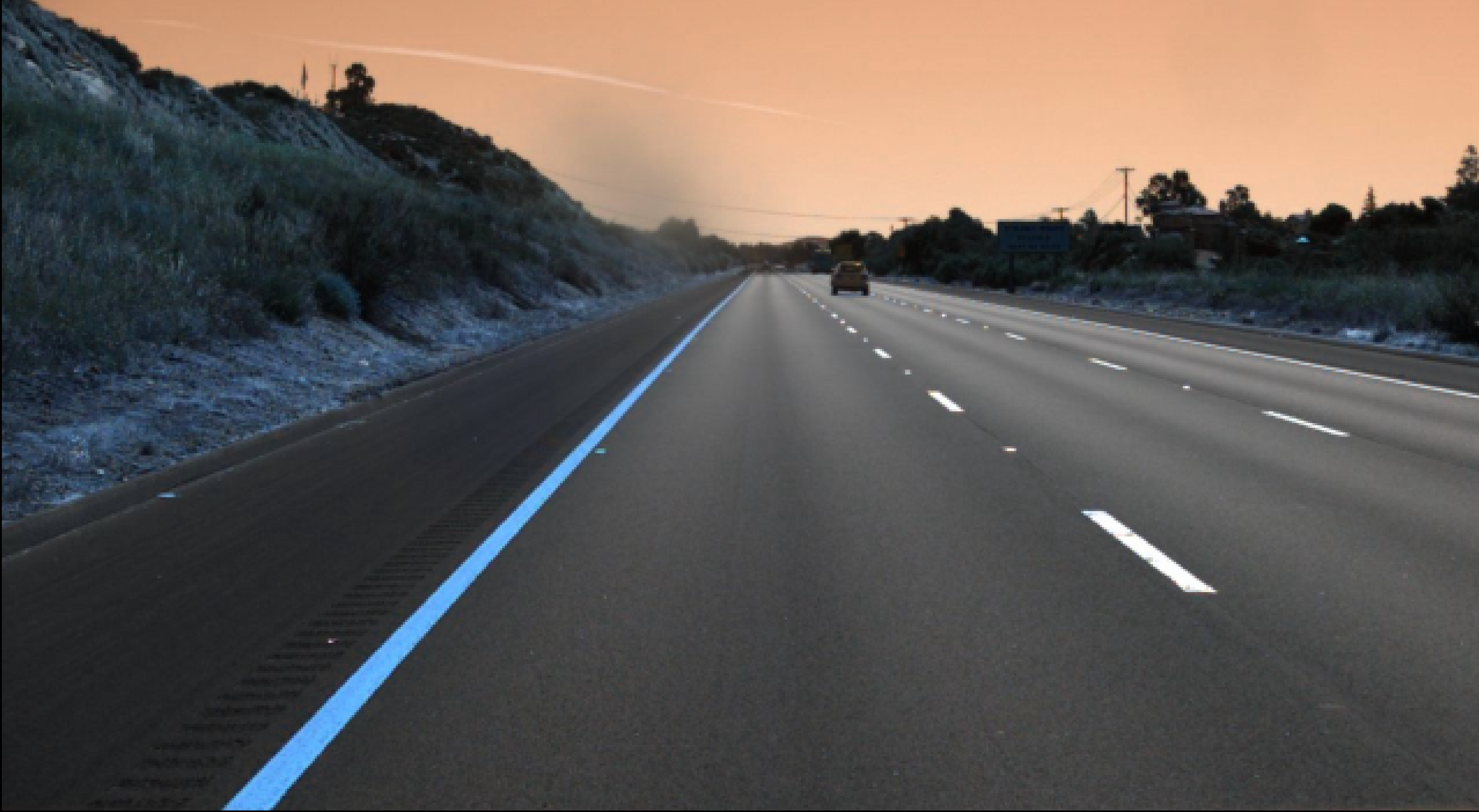}
    \includegraphics[width=2.4cm]{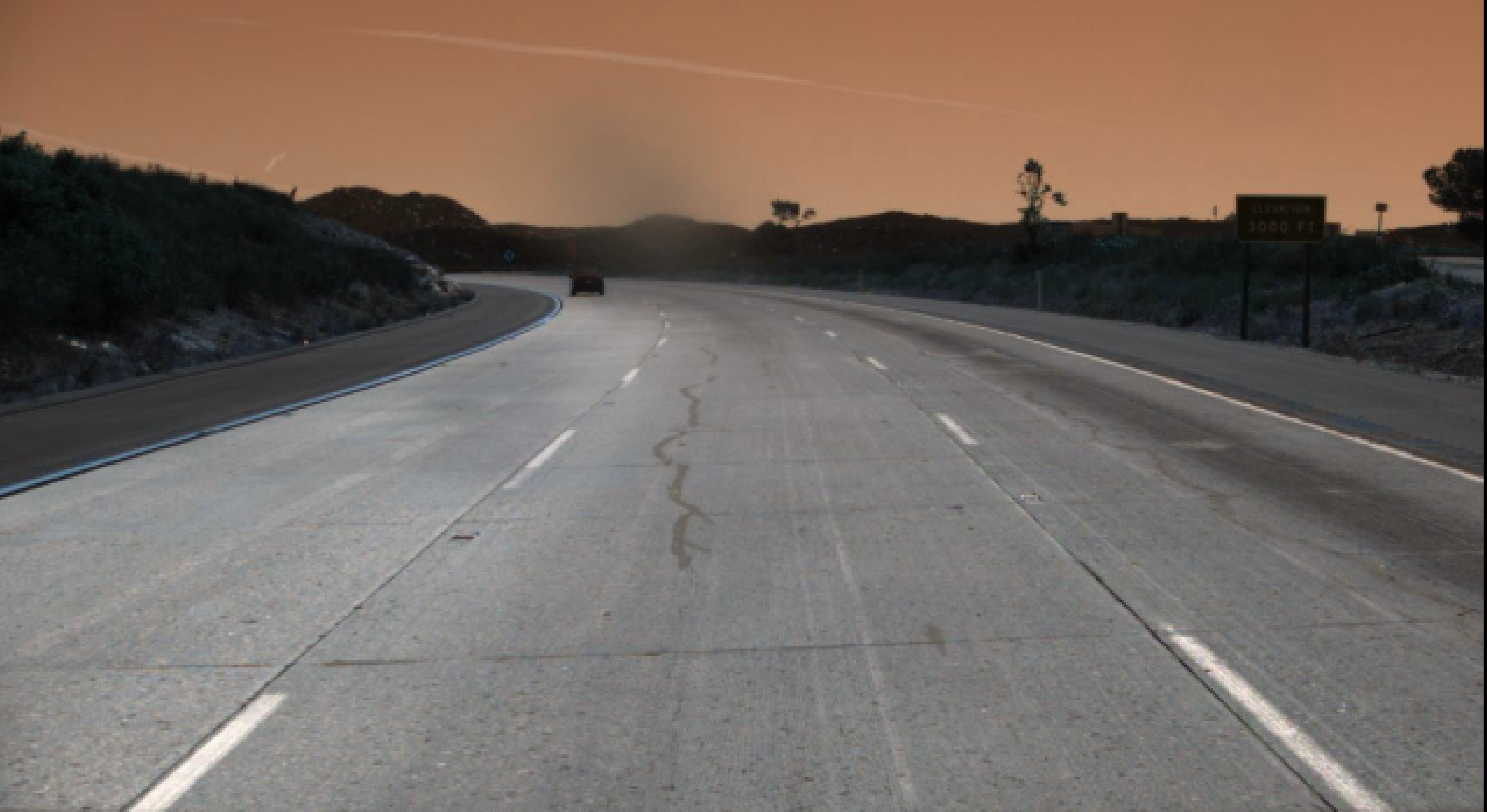} 
    \includegraphics[width=2.4cm]{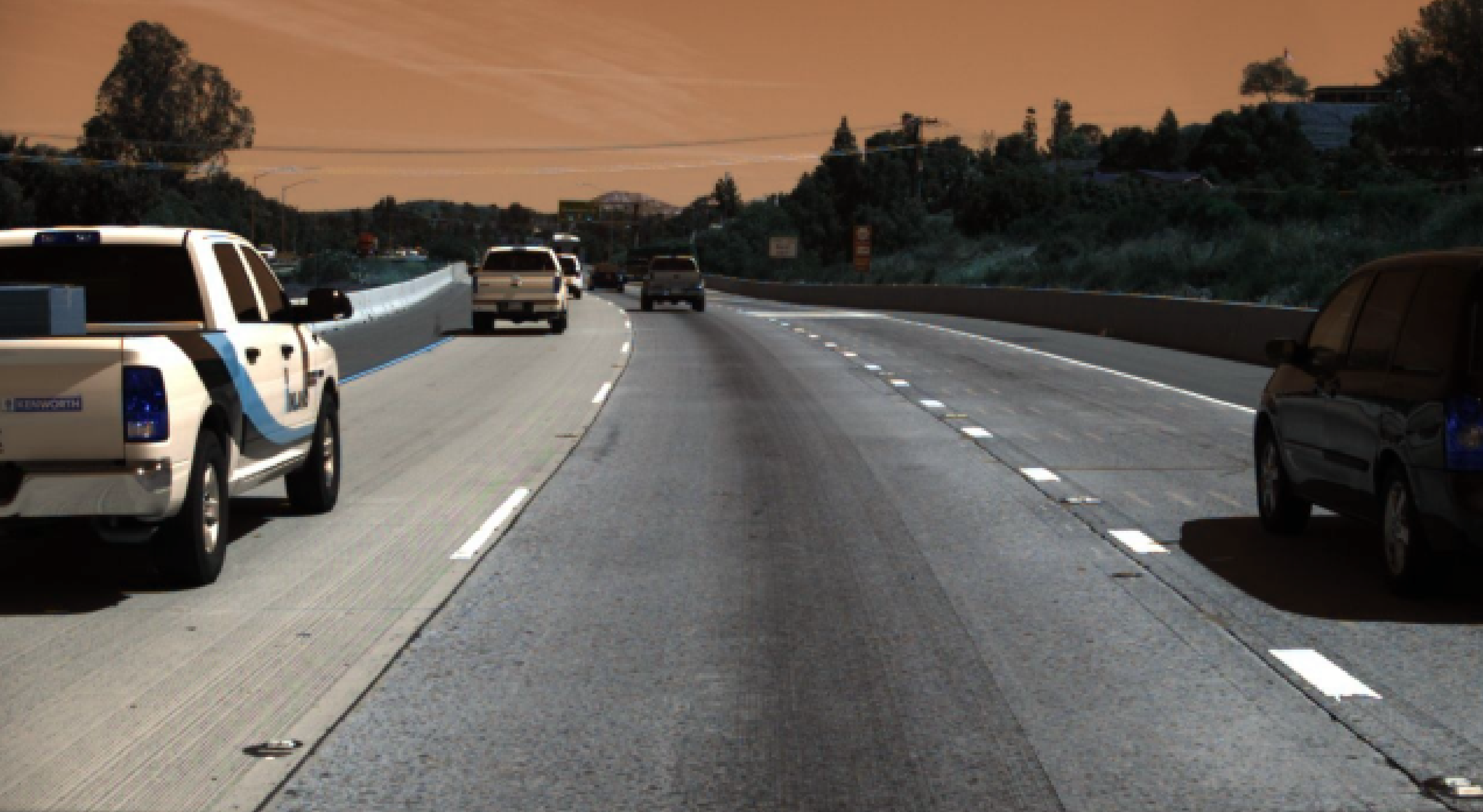}
    \\ [\smallskipamount]
    \includegraphics[width=2.4cm]{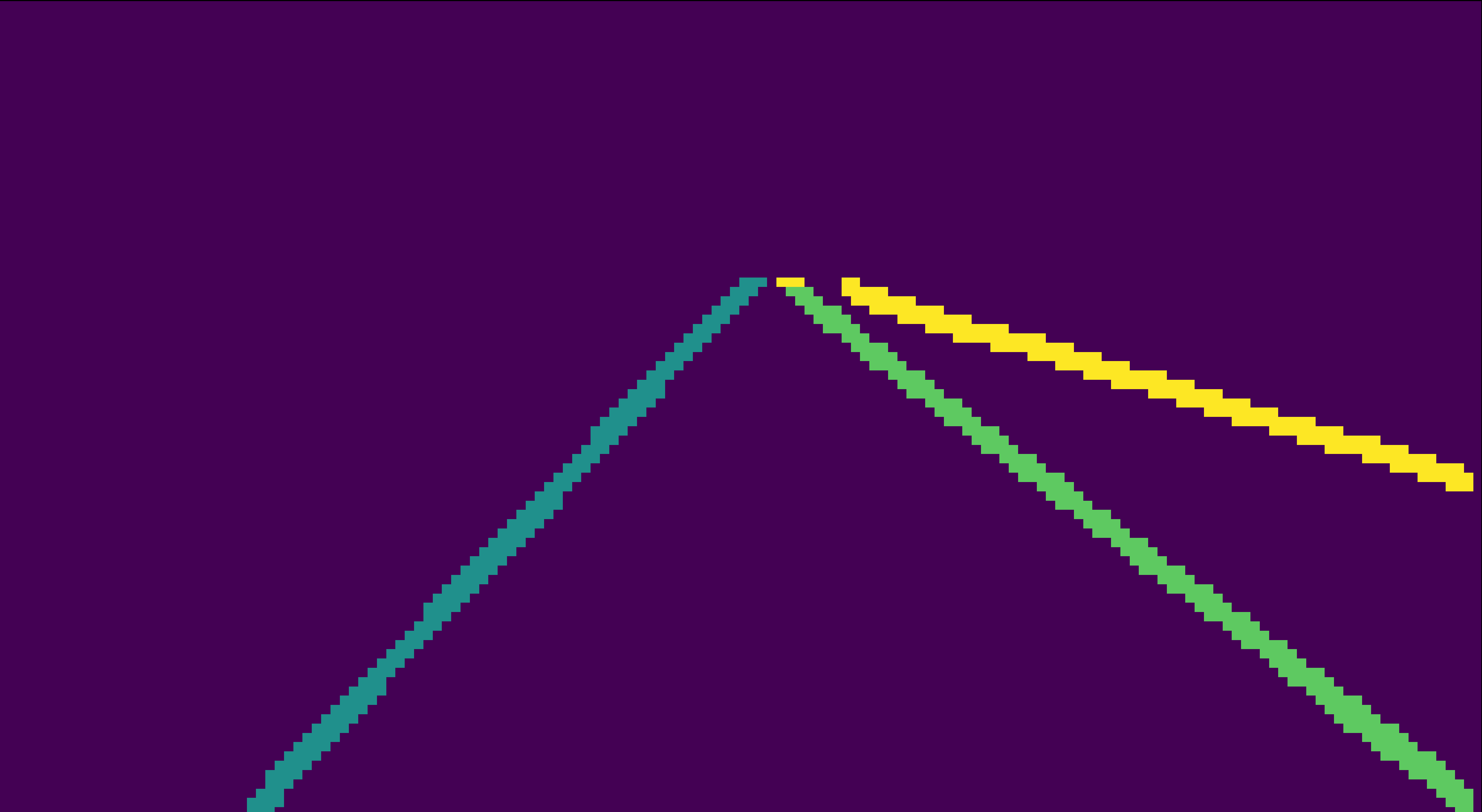}
    \includegraphics[width=2.4cm]{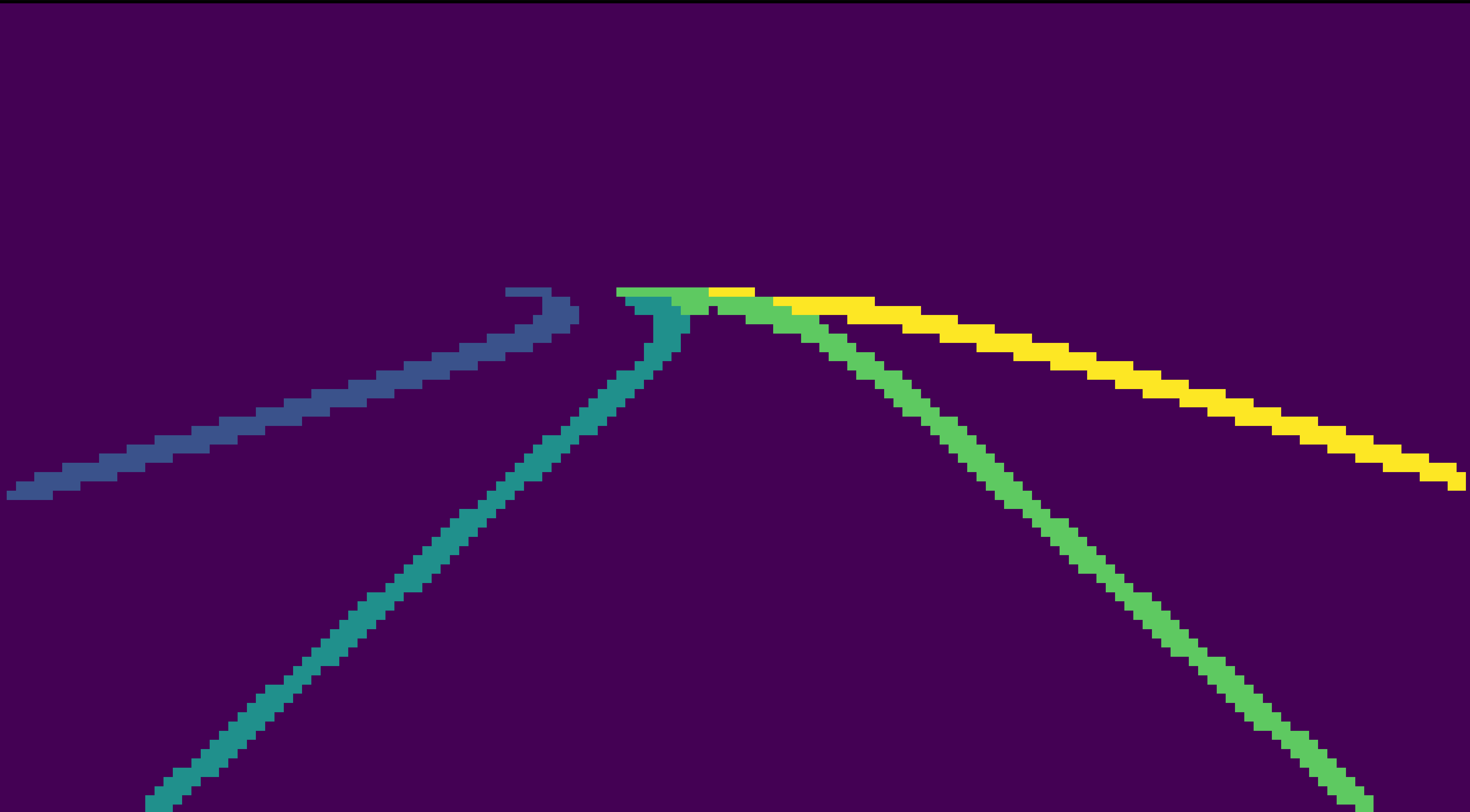} 
    \includegraphics[width=2.4cm]{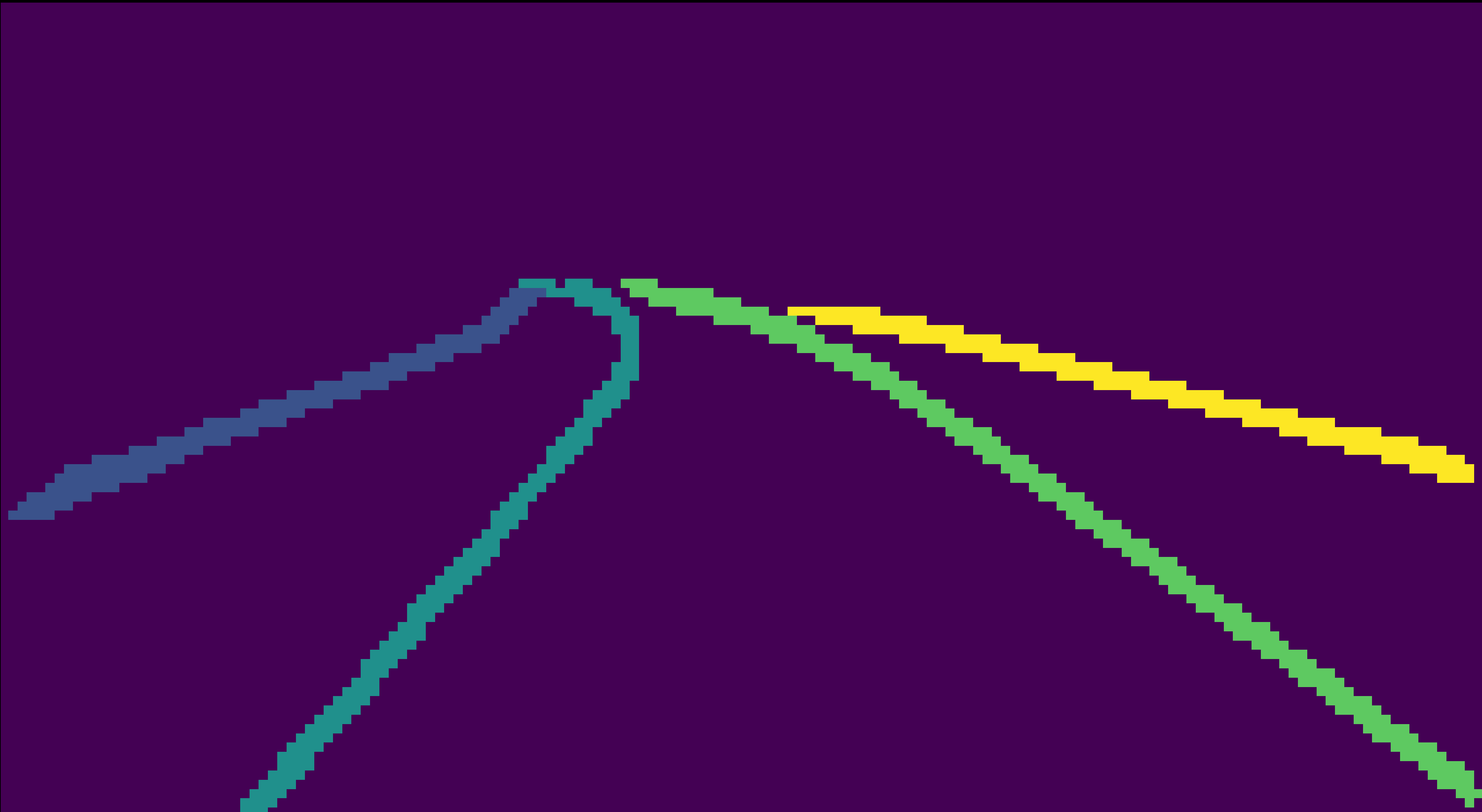}
    \\ [\smallskipamount]
    \includegraphics[width=2.4cm]{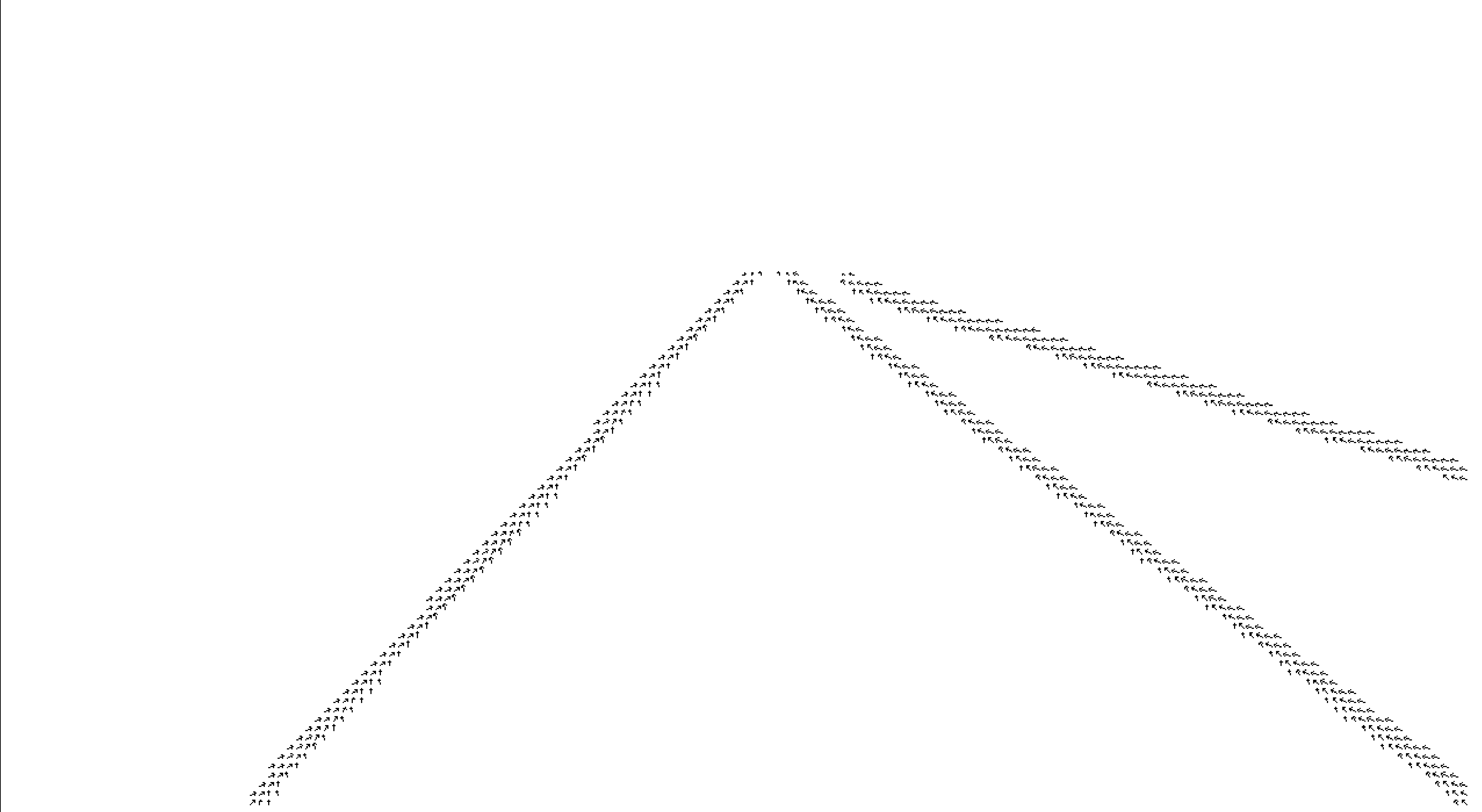}
    \includegraphics[width=2.4cm]{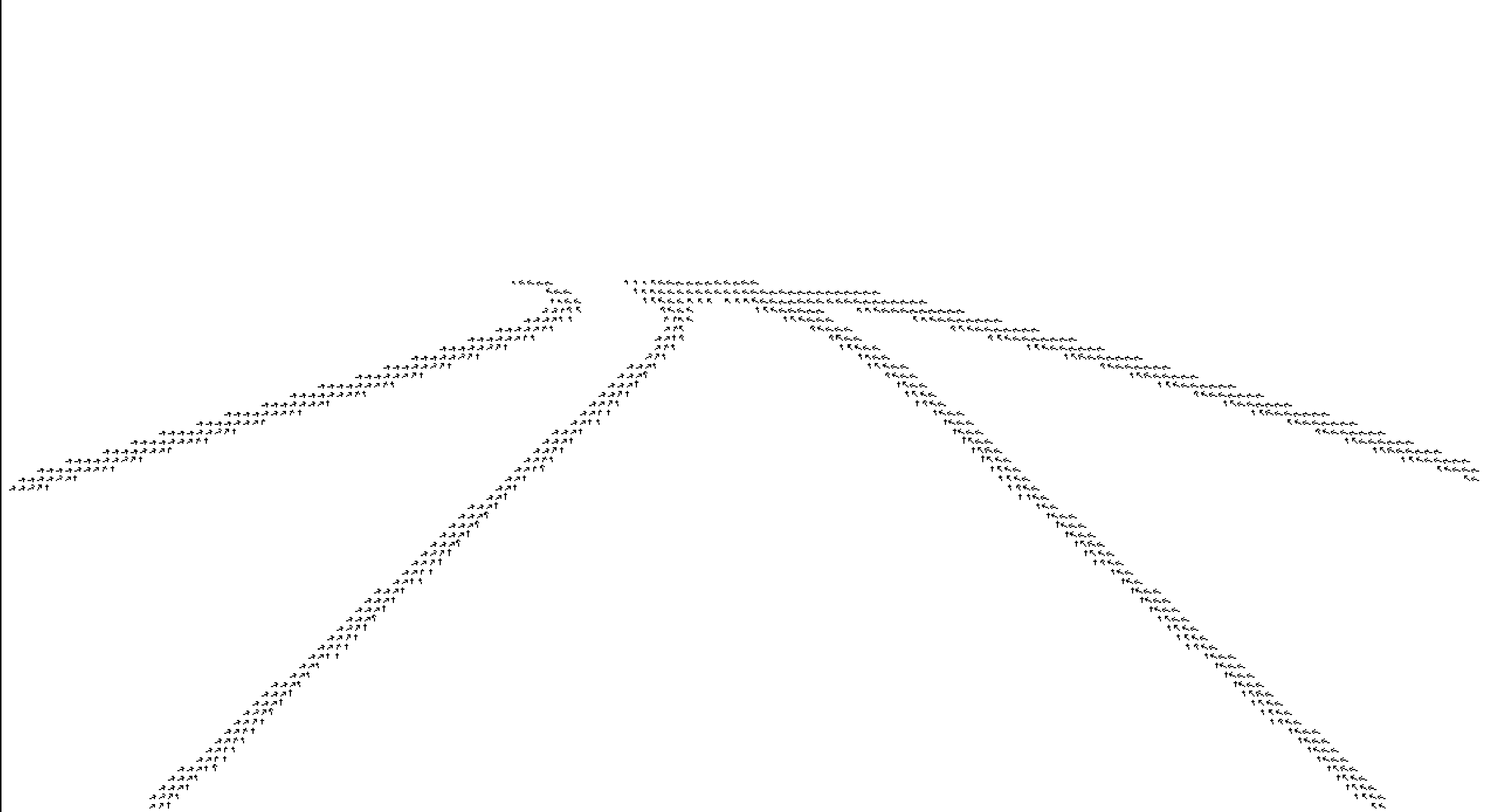} 
    \includegraphics[width=2.4cm]{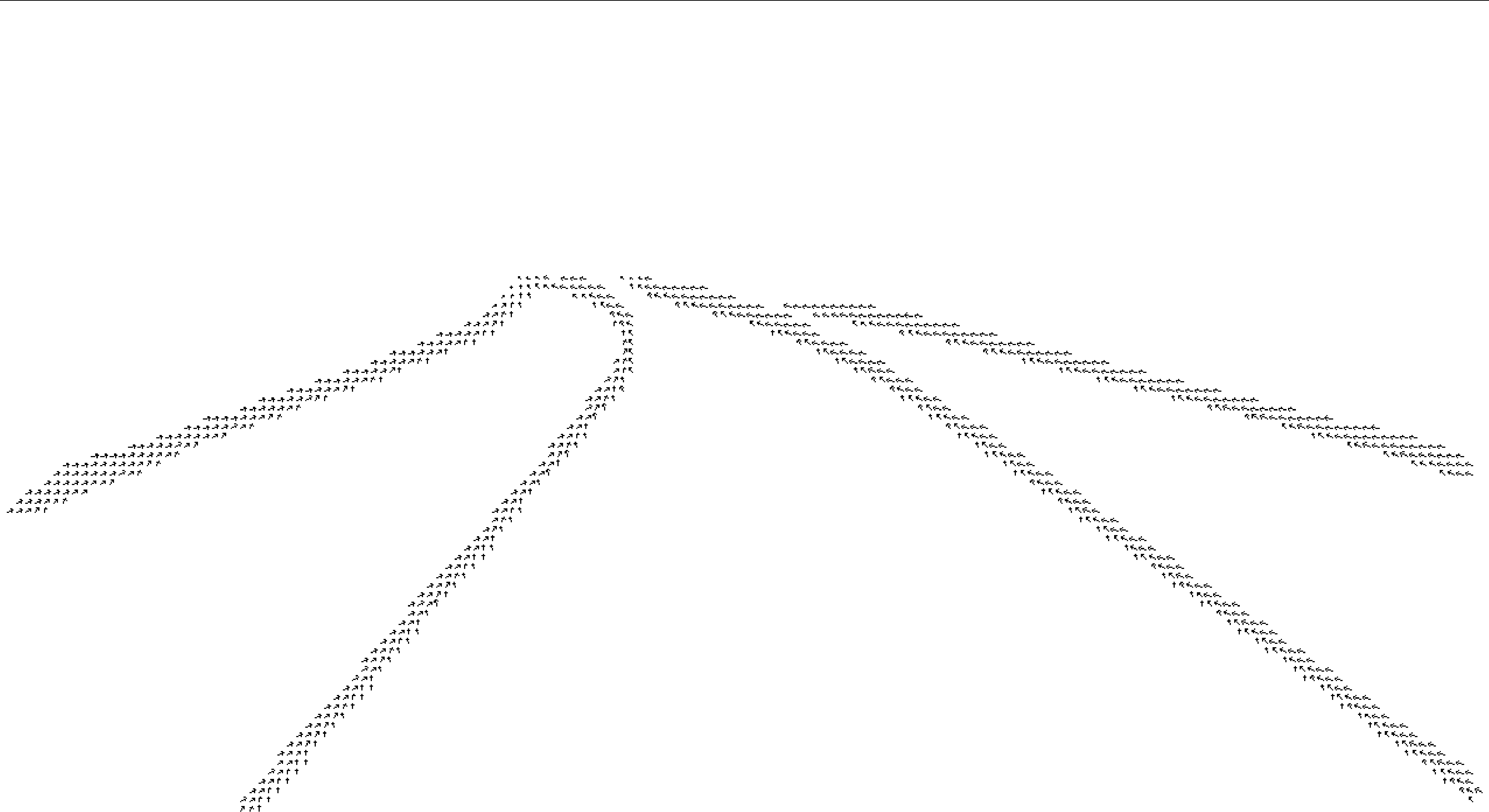}
    \\ [\smallskipamount]
    \includegraphics[width=2.4cm]{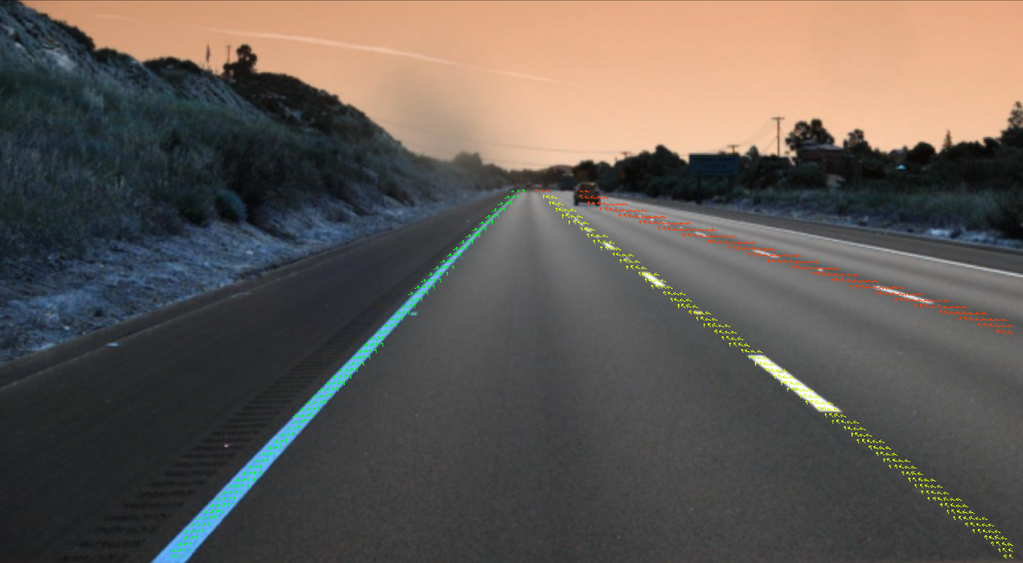}
    \includegraphics[width=2.4cm]{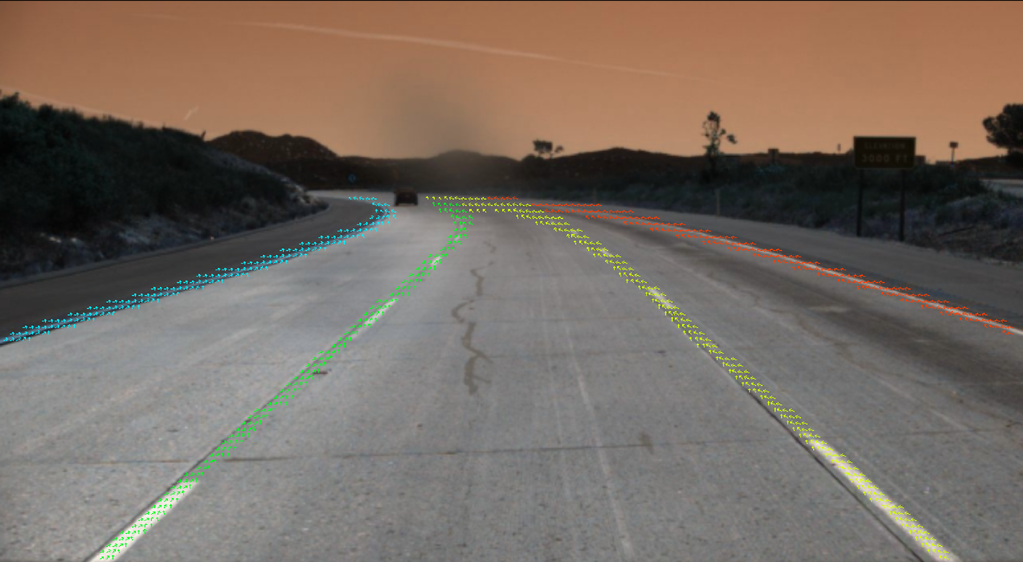} 
    \includegraphics[width=2.4cm]{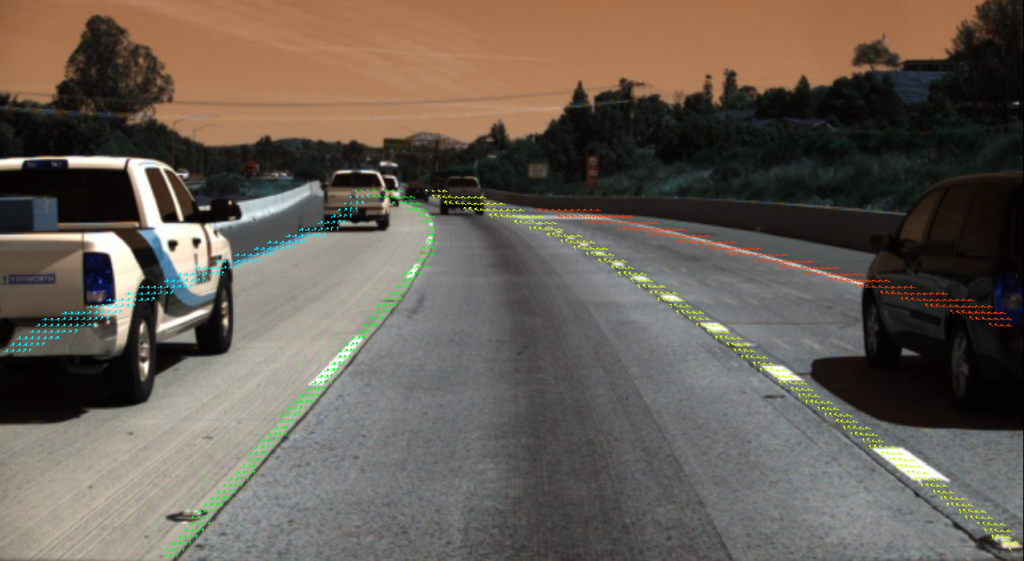}
    \\ [\smallskipamount]
    \caption{Qualitative results on TuSimple dataset.} \label{f2}
\end{figure}

The localization of lane markers with the affinity fields approach has been shown in Fig. \ref{f2}  for the TuSimple dataset. In Fig. \ref{f2}, the first row is the input image, the second one indicates instance segmentation after clustering, the third row shows affinity fields output, and the last one illustrates the final result of the procedure. The proposed method often fails to predict the very end of lanes when an extreme curve or occlusions of lanes exist. The results illustrated in Fig. \ref{f2}determine the high efficiency on highways with curved roads, lanes changing scenarios, and splitting because of highway exits, highlighting our approach’s ability to adjust to the various number of lanes existing on different roads.

In addition, we performed ablation experiments to validate our configuration choices in this study. We performed all ablation studies on the TuSimple dataset. The impact of adjusting various hyper-parameters such as dropout, learning rate, weight decay, and activation functions was investigated. Further, we evaluated the impacts of different layer configurations, including the number of encoding and decoding layers and substituting dilated layers with alternative types of convolution layers such as asymmetric, simple, flattened, and spatial. Despite attempting to improve performance using a stacked structure with local learning, our results fell short of previous findings. Eventually, we selected an optimized architecture with the most efficient hyper-parameter values. This architecture boasts a compact structure, as shown in Table \ref{t4}, with only 0.25 million parameters and 3.14G FLOPs, making it faster than commonly used models such as ResNet family, ERFNet, and ENet while preserving similar accuracy for lane detection. We also applied different types of image noise, including Gaussian and speckle, to assess the robustness of our method during our experiments.
\section{Conclusion}\label{sec5}
In conclusion, this study proposes a novel and efficient CNN architecture for end-to-end lane detection that utilizes semantic segmentation and affinity fields. Our method, influenced by recent advancements in segmentation approaches, employs binary segmentation outputs and affinity fields to perform instance segmentation of lane pixels, allowing for the detection of an inconsistent number of lanes in post-processing. With fewer parameters than other models, our proposed architecture outperforms most of them and achieves a low false positive rate on the TuSimple dataset. Furthermore, the strength of our model is that it has the lowest computational complexity, such as FLOPs and number of parameters, among all existing models. Overall, our work provides a promising solution for lane detection in autonomous driving applications.

\subsection*{Acknowledgments}
The authors declare that they received no funds, grants, or other support during the preparation of this manuscript.
\section*{\large Compliance with ethical standards}

\subsection*{Conflict of interest}
The authors (Seyed Rasoul Hosseini and Mohammad Teshnehlab) declare no conflict of interest.

\subsection*{Ethical approval}
This article contains no studies with human participants or animals performed by any of the authors.

\subsection*{Informed consent}
Informed consent was obtained from all individual participants included in the study.


\bibliography{sn-bibliography}

\begin{thebibliography}{}

\bibitem[Abualsaud et~al., 2021]{b7}
Abualsaud, H., Liu, S., Lu, D.~B., Situ, K., Rangesh, A., and Trivedi, M.~M. (2021).
\newblock Laneaf: Robust multi-lane detection with affinity fields.
\newblock {\em IEEE Robotics and Automation Letters}, 6:7477--7484.

\bibitem[Chetlur et~al., 2014]{b29}
Chetlur, S., Woolley, C., Vandermersch, P., Cohen, J., Tran, J., Catanzaro, B., and Shelhamer, E. (2014).
\newblock cudnn: Efficient primitives for deep learning.
\newblock Preprint at \url{https://doi.org/10.48550/arXiv.1410.0759}.

\bibitem[Ghafoorian et~al., 2018]{b20}
Ghafoorian, M., Nugteren, C., Baka, N., Booij, O., and Hofmann, M. (2018).
\newblock El-gan: Embedding loss driven generative adversarial networks for lane detection.
\newblock In {\em proceedings of the european conference on computer vision (ECCV) Workshops}.

\bibitem[He et~al., 2016a]{b3}
He, B., Ai, R., Yan, Y., and Lang, X. (2016a).
\newblock Accurate and robust lane detection based on dual-view convolutional neutral network.
\newblock In {\em 2016 IEEE Intelligent Vehicles Symposium (IV)}, pages 1041--1046. IEEE.

\bibitem[He et~al., 2015]{b27}
He, K., Zhang, X., Ren, S., and Sun, J. (2015).
\newblock Delving deep into rectifiers: Surpassing human-level performance on imagenet classification.
\newblock In {\em Proceedings of the IEEE international conference on computer vision}, pages 1026--1034.

\bibitem[He et~al., 2016b]{b23}
He, K., Zhang, X., Ren, S., and Sun, J. (2016b).
\newblock Deep residual learning for image recognition.
\newblock In {\em Proceedings of the IEEE conference on computer vision and pattern recognition}, pages 770--778.

\bibitem[Hou, 2019]{b30}
Hou, Y. (2019).
\newblock Agnostic lane detection.
\newblock Preprint at \url{http://arxiv.org/abs/1905.03704}.

\bibitem[Hou et~al., 2019]{b19}
Hou, Y., Ma, Z., Liu, C., and Loy, C.~C. (2019).
\newblock Learning lightweight lane detection cnns by self attention distillation.
\newblock In {\em Proceedings of the IEEE/CVF international conference on computer vision}, pages 1013--1021.

\bibitem[Huval et~al., 2015]{b2}
Huval, B., Wang, T., Tandon, S., Kiske, J., Song, W., Pazhayampallil, J., Andriluka, M., Rajpurkar, P., Migimatsu, T., Cheng-Yue, R., Mujica, F., Coates, A., and Ng, A. (2015).
\newblock An empirical evaluation of deep learning on highway driving.
\newblock Preprint at \url{https://arxiv.org/abs/1504.01716}.

\bibitem[Ioffe and Szegedy, 2015]{b26}
Ioffe, S. and Szegedy, C. (2015).
\newblock Batch normalization: Accelerating deep network training by reducing internal covariate shift.
\newblock In {\em International conference on machine learning}, pages 448--456. pmlr.

\bibitem[Ko et~al., 2021]{b10}
Ko, Y., Lee, Y., Azam, S., Munir, F., Jeon, M., and Pedrycz, W. (2021).
\newblock Key points estimation and point instance segmentation approach for lane detection.
\newblock {\em IEEE Transactions on Intelligent Transportation Systems}, 23:8949--8958.

\bibitem[Li et~al., 2019]{b21}
Li, X., Li, J., Hu, X., and Yang, J. (2019).
\newblock Line-cnn: End-to-end traffic line detection with line proposal uni.
\newblock {\em IEEE Transactions on Intelligent Transportation Systems}, 21:248--258.

\bibitem[Liu et~al., 2021a]{b12}
Liu, L., Chen, X., Zhu, S., and Tan, P. (2021a).
\newblock Condlanenet: a top-to-down lane detection framework based on conditional convolution.
\newblock In {\em Proceedings of the IEEE/CVF international conference on computer vision}, pages 3773--3782.

\bibitem[Liu et~al., 2021b]{b15}
Liu, R., Yuan, Z., Liu, T., and Xiong, Z. (2021b).
\newblock End-to-end lane shape prediction with transformers.
\newblock In {\em Proceedings of the IEEE/CVF winter conference on applications of computer vision}, pages 3694--3702.

\bibitem[Neven et~al., 2018]{b5}
Neven, D., De~Brabandere, B., Georgoulis, S., Proesmans, M., and Van~Gool, L. (2018).
\newblock Towards end-to-end lane detection: an instance segmentation approach.
\newblock In {\em 2018 IEEE intelligent vehicles symposium (IV)}, pages 286--291. IEEE.

\bibitem[Pan et~al., 2018]{b4}
Pan, X., Shi, J., Luo, P., Wang, X., and Tang, X. (2018).
\newblock Spatial as deep: Spatial cnn for traffic scene understanding.
\newblock In {\em Proceedings of the AAAI Conference on Artificial Intelligence}, volume~32.

\bibitem[Paszke et~al., 2016]{b24}
Paszke, A., Chaurasia, A., Kim, S., and Culurciello, E. (2016).
\newblock Enet: A deep neural network architecture for real-time semantic segmentation.
\newblock {\em arXiv preprint arXiv:1606.02147}.

\bibitem[Qin et~al., 2020]{b13}
Qin, Z., Wang, H., and Li, X. (2020).
\newblock Ultra fast structure-aware deep lane detection.
\newblock In {\em Computer Vision--ECCV 2020: 16th European Conference, Glasgow, UK, August 23--28, 2020, Proceedings, Part XXIV 16}, pages 276--291. Springer.

\bibitem[Qu et~al., 2021]{b11}
Qu, Z., Jin, H., Zhou, Y., Yang, Z., and Zhang, W. (2021).
\newblock Focus on local: Detecting lane marker from bottom up via key point.
\newblock In {\em Proceedings of the IEEE/CVF Conference on Computer Vision and Pattern Recognition}, pages 14122--14130.

\bibitem[Rezatofighi et~al., 2019]{b22}
Rezatofighi, H., Tsoi, N., Gwak, J.~Y., Sadeghian, A., Reid, I., and Savarese, S. (2019).
\newblock Generalized intersection over union: A metric and a loss for bounding box regression.
\newblock In {\em Proceedings of the IEEE/CVF conference on computer vision and pattern recognition}, pages 658--666.

\bibitem[Romera et~al., 2017]{b25}
Romera, E., Alvarez, J.~M., Bergasa, L.~M., and Arroyo, R. (2017).
\newblock Erfnet: Efficient residual factorized convnet for real-time semantic segmentation.
\newblock {\em IEEE Transactions on Intelligent Transportation Systems}, 19:263--272.

\bibitem[Tabelini et~al., 2021a]{b17}
Tabelini, L., Berriel, R., Paixao, T.~M., Badue, C., De~Souza, A.~F., and Oliveira-Santos, T. (2021a).
\newblock Keep your eyes on the lane: Real-time attention-guided lane detection.
\newblock In {\em Proceedings of the IEEE/CVF conference on computer vision and pattern recognition}, pages 294--302.

\bibitem[Tabelini et~al., 2021b]{b16}
Tabelini, L., Berriel, R., Paixao, T.~M., Badue, C., De~Souza, A.~F., and Oliveira-Santos, T. (2021b).
\newblock Polylanenet: Lane estimation via deep polynomial regression.
\newblock In {\em 2020 25th International Conference on Pattern Recognition (ICPR)}, pages 6150--6156. IEEE.

\bibitem[Tang et~al., 2021]{b1}
Tang, J., Li, S., and Liu, P. (2021).
\newblock A review of lane detection methods based on deep learning.
\newblock {\em Pattern Recognition}, 111.

\bibitem[TuSimple, 2017]{b18}
TuSimple (2017).
\newblock Tusimple lane detection benchmark.
\newblock figshare \url{https://github.com/Tusimple/Tusimple-benchmark}.

\bibitem[Wang et~al., 2022]{b6}
Wang, J., Ma, Y., Huang, S., Hui, T., Wang, F., Qian, C., and Zhang, T. (2022).
\newblock A keypoint-based global association network for lane detection.
\newblock In {\em Proceedings of the IEEE/CVF Conference on Computer Vision and Pattern Recognition}, pages 1392--1401.

\bibitem[Wang et~al., 2018]{b8}
Wang, Z., Ren, W., and Qiu, Q. (2018).
\newblock Lanenet: Real-time lane detection networks for autonomous driving.
\newblock Preprint at \url{https://arxiv.org/abs/1807.01726}.

\bibitem[Yoo et~al., 2020]{b14}
Yoo, S., Lee, H.~S., Myeong, H., Yun, S., Park, H., Cho, J., and Kim, D.~H. (2020).
\newblock End-to-end lane marker detection via row-wise classification.
\newblock In {\em Proceedings of the IEEE/CVF Conference on Computer Vision and Pattern Recognition Workshops}, pages 1006--1007.

\bibitem[Zheng et~al., 2021]{b9}
Zheng, T., Fang, H., Zhang, Y., Tang, W., Yang, Z., Liu, H., and Cai, D. (2021).
\newblock Resa: Recurrent feature-shift aggregator for lane detection.
\newblock In {\em Proceedings of the AAAI Conference on Artificial Intelligence}, volume~35, pages 3547--3554.

\bibitem[Zheng et~al., 2022]{b31}
Zheng, T., Huang, Y., Liu, Y., Tang, W., Yang, Z., Cai, D., and He, X. (2022).
\newblock Clrnet: Cross layer refinement network for lane detection.
\newblock In {\em Proceedings of the IEEE/CVF conference on computer vision and pattern recognition}, pages 898--907.

\end{thebibliography}
\bibliographystyle{apalike}

\end{document}